\documentclass[10pt,journal,compsoc]{IEEEtran}
\usepackage[left=1.5cm,right=1.5cm,top=2cm,bottom=2cm]{geometry}
\usepackage{ragged2e}
\justifying

\ifCLASSOPTIONcompsoc
   \usepackage[nocompress]{cite}
\else
    \usepackage{cite}
\fi

\ifCLASSINFOpdf
 
\else
 
\fi
\usepackage{multirow}
\usepackage{graphicx}
\usepackage{xcolor}
\usepackage{tabularx}

\usepackage{authblk}

\begin{document}

\title{Building an Effective Email Spam Classification Model with spaCy}
\author{\IEEEauthorblockN{Kazem Taghandiki\IEEEauthorrefmark{1}
                        }
	\IEEEauthorblockA{
	 \\
		\IEEEauthorrefmark{1}Department of Computer Engineering, Technical and Vocational University (TVU), Tehran, Iran
 \\ktaghandiki@tvu.ac.ir
 \\
  		                    }
							}
\IEEEtitleabstractindextext{%
\begin{abstract}
\justifying
\textcolor{black}{Today, people use email services such as Gmail, Outlook, AOL Mail, etc. to communicate with each other as quickly as possible to send information and official letters. Spam or junk mail is a major challenge to this type of communication, usually sent by botnets with the aim of advertising, harming and stealing information in bulk to different people. Receiving unwanted spam emails on a daily basis fills up the inbox folder. Therefore, spam detection is a fundamental challenge, so far many works have been done to detect spam using clustering and text categorisation methods. In this article, the author has used the spaCy natural language processing library and 3 machine learning (ML) algorithms Naive Bayes (NB), Decision Tree C45 and Multilayer Perceptron (MLP) in the Python programming language to detect spam emails collected from the Gmail service. Observations show the accuracy rate (96$\%$) of the Multilayer Perceptron (MLP) algorithm in spam detection. }
\end{abstract}
\begin{IEEEkeywords}
Machine Learning, Naive Bayes, Decision Tree, Natural Language Processing, Multi-Layer Perceptron, Spam Email
\end{IEEEkeywords}}

\maketitle

\IEEEdisplaynontitleabstractindextext

\IEEEpeerreviewmaketitle

\section{Introduction}
The Internet has become an essential part of people's lives. On the Internet, people send and receive messages through email services such as Gmail, Outlook, AOL Mail, etc. Assuming that one or more advertising messages are sent to a person every day, if this person is not aware of spam, he or she will easily fall into the trap of this type of message\cite{a1}. The main purpose of spam is to advertise, damage the recipient's system and steal important information from the recipient of a message \cite{a18,b1}.\\
Spam detection systems focus more on the textual content of a message. These types of systems first receive a textual dataset (message set) as input and, after performing pre-processing operations such as removing stop words, normalizing, stemming, etc., the message is processed using various natural language processing and machine learning algorithms\cite{a3}. Usually, people who want to send spam messages collect a collection of users' emails from blogs, forums, then write a targeted letter (with the aim of advertising or stealing) for a collection of emails. Collected are sent. Upon seeing this message, the recipient of the email may read it and send their important information (bank card details, password, etc \cite{b2,b3}.) in response. Users of the United Nations, for example, have been victims of this type of attack\cite{a4}.\\
If users of email servers recognise a message as spam, it is better not to click on any links or attachments\cite{a13}. Spammers sometimes add unsubscribe or unsubscribe links to verify that your email address is active; these types of links usually steal information, so users should not click on them\cite{a13,a4}.\\
Spam messages are difficult to stop because they can be sent through botnets, which are networks of pre-infected computers that make it difficult to track and stop the original spam \cite{a12}.\\
In the proposed approach, a set of useful and spam messages is first collected through Gmail, then text pre-processing is performed on the textual content of the messages using the spaCy tool. Finally, by using the machine learning algorithms Naive Bayes (NB), Decision Tree C45 and Multilayer Perceptron (MLP) in Python programming language, the spam detection process has been done.
\subsection{What is spaCy}
spaCy \cite{a9} is a free and open-source library for natural language processing (NLP) in Python. It provides various NLP capabilities such as Named Entity Recognition (NER), Part-Of-Speech (POS) tagging, dependency parsing, and word vectors. spaCy is designed to make it easy to build systems for information extraction, text classification, and other NLP tasks \cite{b5,c4}.It can handle large volumes of text data and is widely used in data science and machine learning.
\subsection{Naive Bayes (NB)}
Naive Bayes is a simple and fast classification algorithm based on Bayes' theorem\cite{a25,b6}. It is used for assigning class labels to problem instances represented as vectors of feature values. Naive Bayes classifiers are a collection of classification algorithms that can be used for various tasks such as text classification, spam filtering, and sentiment analysis. The algorithm assumes that the features are independent, making it computationally efficient and easy to implement. Naive Bayes models are often suitable for high-dimensional datasets and can provide good results with limited training data \cite{b4,c1}.
\subsection{Decision Tree C45\cite{a24}}
A decision tree is a non-parametric supervised learning algorithm used for both classification and regression tasks. It is a decision support tool that uses a tree-like model of decisions and their possible consequences, including chance event outcomes. A decision tree is a flowchart-like tree structure consisting of nodes, branches, internal nodes, and leaf nodes. The root node represents the entire population or sample. The internal nodes represent features or attributes of the data, while the branches represent the possible values of these features. The leaf nodes represent the outcome or target variable.
\subsection{Multilayer Perceptron (MLP)}
A Multilayer Perceptron (MLP) \cite{a26} is a fully connected class of feedforward Artificial Neural Network (ANN) \cite{a7,c2}. It consists of three types of layers: the input layer, output layer, and one or more hidden layers with many neurons stacked together.In contrast to the Perceptron, where the neuron must have an activation function that imposes a threshold like ReLU or sigmoid, neurons in an MLP can use any arbitrary activation function. An MLP is trained using backpropagation, which adjusts the weights of the connections between neurons to minimize the error between predicted and actual outputs.\\
Therefore, in the continuation of this article and in section 2, a series of works done in the field of spam detection have been examined, then in section 3, the proposed approach is presented in full. Section 4 examines the results obtained from the proposed approach, and finally, Section 5 discusses the conclusions of the proposed approach.  
\section{Related Work}
A lot of work has been done to detect spam, most of it using natural language processing tools such as NLTK and SVM machine learning algorithms. However, in recent years, various machine learning and deep learning algorithms\cite{a14,a15,b8,a16,a17,b9} have also been used to detect spam.\\
Mohammad et al\cite{a5} presented a 3-layer processing system for spam detection. In the first layer of the proposed system, emails are fed into the system as an input dataset. In the second layer, various natural language processing libraries such as NLTK are used to perform preprocessing and feature extraction. Finally, in the third layer, the Naive Bayes machine learning algorithm is also used to detect spam or usefulness of incoming emails.\\
Marsono et al\cite{a6} have presented a hardware architecture with the purpose of preventive management to detect spam or useful emails. The architecture provided by them received 117 million emails as input every second and after performing various text pre-processing operations, it was used to detect spam or usefulness of emails using SVM machine learning algorithm.\\
Tang et al\cite{a7} presented an approach based on assigning a trust value to an IP address. By extracting IPs from a collection of spam emails, they created a dataset of IPs and trust values. They then used SVM and Random Forest machine learning algorithms to detect unreliable IPs and finally spam. The results showed better and faster accuracy of the SVM algorithm than the Random Forest algorithm.\\
Yoo et al\cite{a8} presented a spam detection approach based on user group formation. In this approach, by forming and creating different groups of people, the user trains the system to recognise the emails received from the created groups as useful emails and the rest of the emails as spam. This approach does not process the textual content of the message or email.
The proposed approach is implemented below. In the proposed approach, a set of useful and spam messages is first collected through Gmail, then text preprocessing is performed on the textual content of the messages using the spaCy tool. Finally, by using the machine learning algorithms Naive Bayes (NB), Decision Tree C45 and Multilayer Perceptron (MLP) \cite{b7} in Python programming language, the spam detection process has been done.
\section{The proposed approach}
To present the proposed approach, the author has used a 3-step process. Figure 1 shows the process of the proposed approach.
\begin{figure}
    \centering
    \includegraphics[width=7.5cm,height=1.5cm]{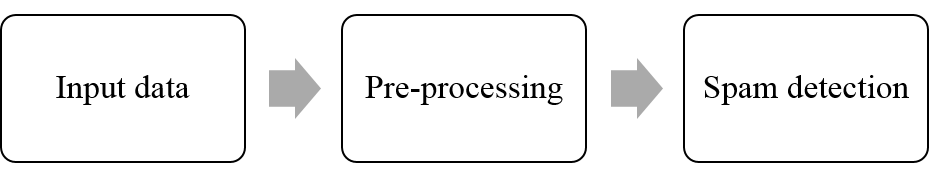}
    \caption{Implementation process of the proposed approach}
    \label{fig:life}
\end{figure}
As shown in Figure 1, it is clear that in the first step, 1500 messages or emails are collected from the Gmail service and provided as input data to the second step of the proposed system process. In the second step, various pre-processing operations such as removal of stop words, removal of numbers, normalisation and stemming are performed on the data set collected from the previous step using the spaCy tool\cite{a9}. Finally, in the third step, three algorithms Naive Bayes (NB), Decision Tree C45 and Multilayer Perceptron (MLP) are used simultaneously to train the model and evaluate the models in spam detection \cite{b10}.
\subsection{Input data}
In this step, 750 spam emails (from the spam folder) and 750 useful emails (from the inbox folder), a total of 1500 emails, are extracted and collected as a primary dataset from the Gmail account. The main reason for the fair division of the input dataset into two parts, useful emails and spam, is to balance the results of the models built in the third step.
\subsection{Pre-processing}
In this step, using the spaCy tool, various pre-processing processes such as tokenization, removal of stop words, removal of numbers, normalization and stemming are performed on the input data from the first step. The purpose of pre-processing is to improve the quality of input data so that the process of detecting spam and useful messages can be done with better accuracy in the third step. For example, Figure 2 shows the tokenization pre-processing process on the content of an email.
\begin{figure}
    \centering
    \includegraphics[width=9cm,height=3cm]{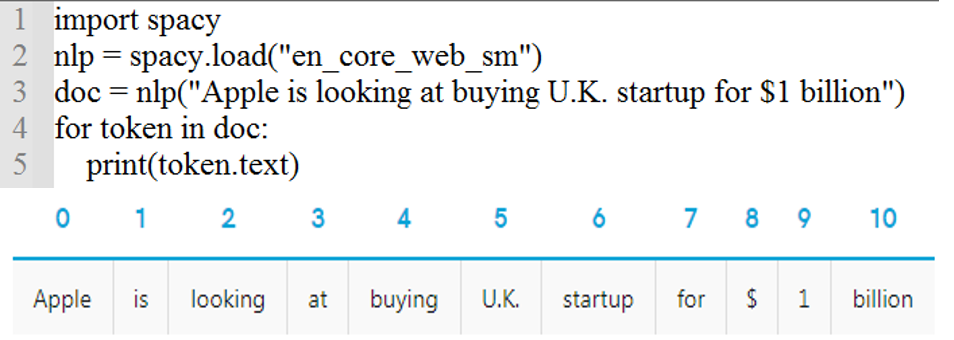}
    \caption{Tokenisation pre-processing}
    \label{fig:life}
\end{figure}
As shown in Figure 2, the text of the message contains 10 words (tokens). Figure 3 shows the pre-processing of tokenization, stemming and stopword detection on the content of the same email. \\
As shown in Figure 3, the TEXT column shows the words of an email message, the LEMMA column shows the structural and lexical root of each word in the TEXT column, and the STOP column shows whether a word is a stop word.
\begin{figure}
    \centering
    \includegraphics[width=9cm,height=9cm]{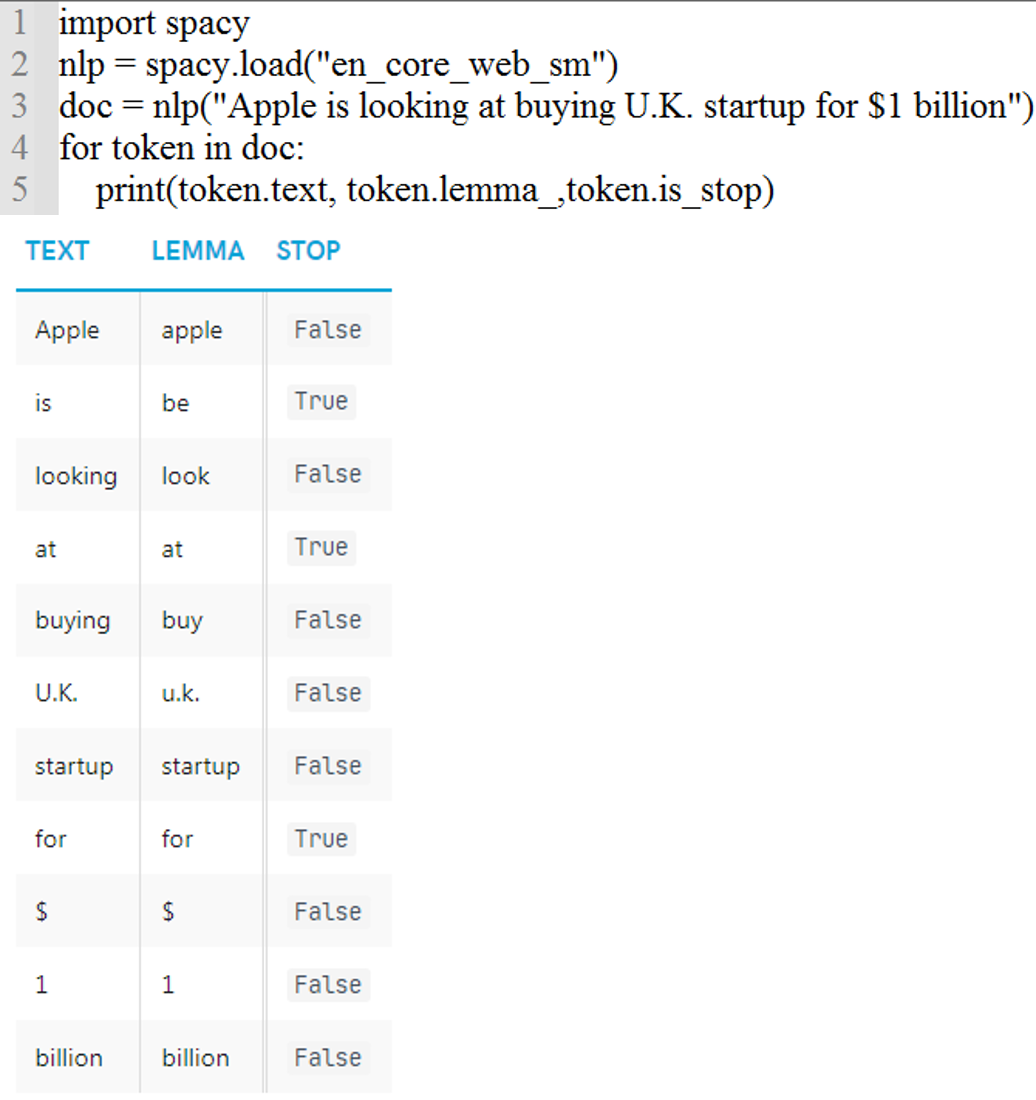}
    \caption{pre-processing of tokenization, stemming and stop-word detection}
    \label{fig:life}
\end{figure}
\subsection{Spam detection}
Once the various pre-processing operations (second step) have been performed on the input data and the quality data has been obtained, different machine learning algorithms can be used to train and evaluate the results.\\
Therefore, in this step, 75$\%$ of the input data (1125 messages) are provided as training data to 3 algorithms Naive Bayes (NB), Decision Tree C45 and Multilayer Perceptron (MLP) to build 3 trained models. Finally, 3 trained models are evaluated on the remaining 25$\%$ of the input data (375 messages) to perform the process of spam detection and email usefulness to obtain 4 evaluation parameters: accuracy, precision, recall and f1-score.
\section{observations}
Figures 5, 4 and 6 show the confusion matrices of each of the machine learning models Naive Bayes (NB), Decision Tree C45 and Multilayer Perceptron (MLP) respectively, according to the real labels of the incoming emails and the predicted labels. It shows by models. TRUE label means useful and False means an email is spam.
\begin{figure}
    \centering
    \includegraphics[width=9cm,height=2cm]{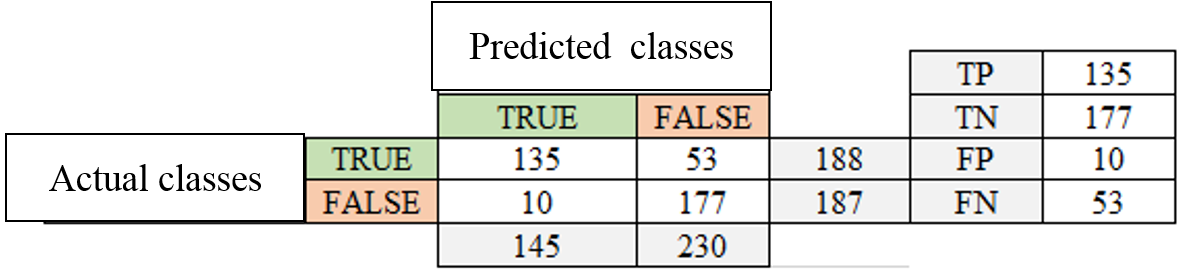}
    \caption{Confusion matrix extraction from the final Naive Bayes (NB) model}
    \label{fig:life}
\end{figure}
\begin{figure}
    \centering
    \includegraphics[width=9cm,height=2cm]{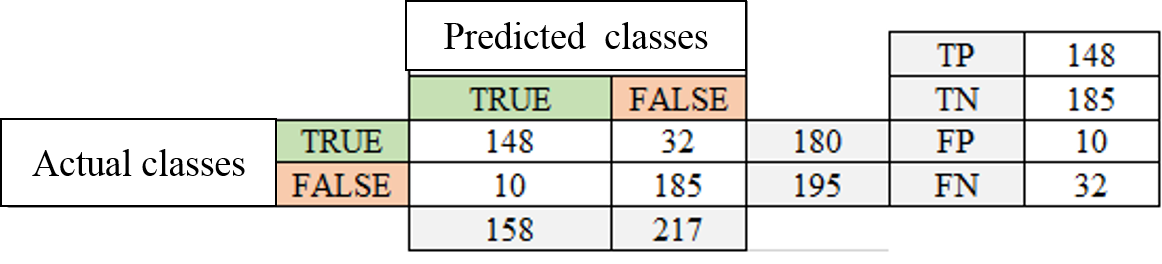}
    \caption{Confusion matrix extraction from the final Decision Tree C45 model}
    \label{fig:life}
\end{figure}
\begin{figure}
    \centering
    \includegraphics[width=9cm,height=2cm]{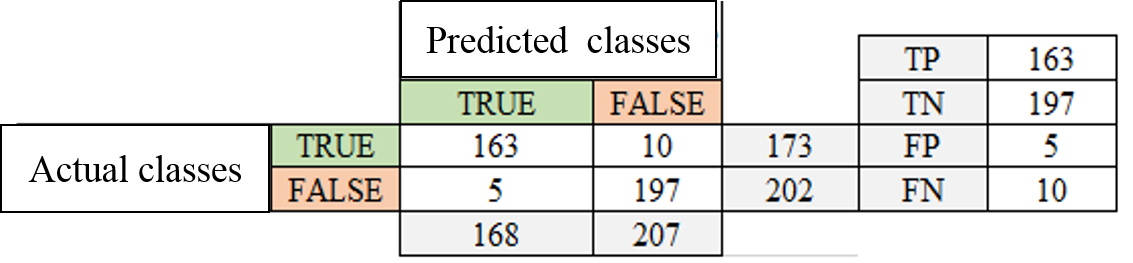}
    \caption{Confusion matrix extraction from the final Multilayer Perceptron (MLP) model}
    \label{fig:life}
\end{figure}
Now, according to the values of the variables TP, FP, TN and FN of the confusion matrices of Figures 5, 4 and 6, it is easy to obtain the evaluation criteria of the machine learning algorithms.\\
Figures 7, 8 and 9 show the statistical results of the accuracy, recall, precision, f1-score evaluation criteria of the constructed models Naive Bayes (NB), Decision Tree C45 and Multilayer Perceptron (MLP).
\begin{figure}
    \centering
    \includegraphics[width=9cm,height=5cm]{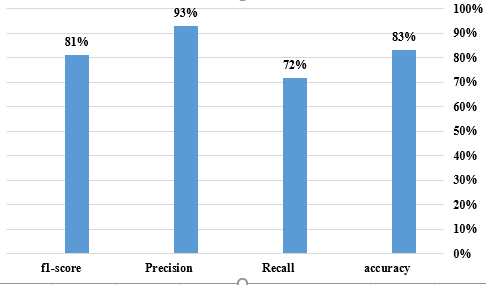}
    \caption{Statistical results of the accuracy, recall, precision and f1 scores of the final Naive Bayes (NB) model.}
    \label{fig:life}
\end{figure}
\begin{figure}
    \centering
    \includegraphics[width=9cm,height=5cm]{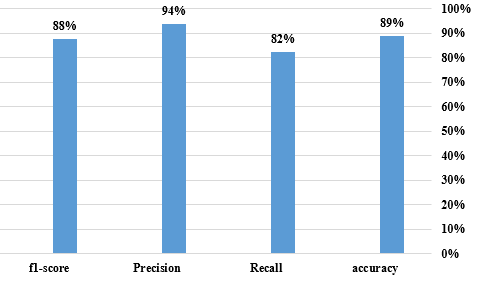}
    \caption{Statistical results of the accuracy, recall, precision and f1 scores of the final Decision Tree C45 model.}
    \label{fig:life}
\end{figure}
\begin{figure}
    \centering
    \includegraphics[width=9cm,height=5cm]{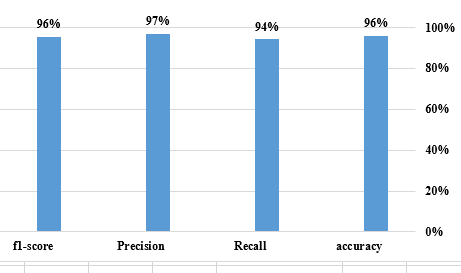}
    \caption{Statistical results of the accuracy, recall, precision and f1 scores of the final Multilayer Perceptron (MLP) model.}
    \label{fig:life}
\end{figure}
As can be seen from Figures 8, 7 and 9, the model obtained from the Multilayer Perceptron (MLP) algorithm has accuracy (96$\%$), recall (94$\%$), precision (97$\%$) and f1-score (96$\%$). Although the recall of the model obtained by the Decision Tree C45 algorithm is higher than that of the Naive Bayes (NB) and Multilayer Perceptron (MLP) algorithms.\\
Certainly, the preprocessing carried out with the spaCy natural language processing library played a very important role in the criteria obtained.
\section{Conclusion}
Spam or junk emails are a fundamental challenge that are sent to people's email accounts in bulk for the purpose of advertising, harming and stealing information and filling their inbox folders. In this article, the author first collected and extracted 1500 useful and spam emails from the Gmail service, then using the natural language processing library spaCy, he applied various text pre-processing operations on the text content of the emails, finally using 3 algorithms Machine Learning Naive Bayes (NB), Decision Tree C45 and Multilayer Perceptron (MLP) in Python programming language was used to train and recognise spam emails collected from the Gmail service. Observations show the accuracy (96$\%$), recall (94$\%$) and precision (97$\%$) of the proposed approach in detecting spam emails.
\section{Future works}
The use of machine learning techniques has been successful in detecting and filtering spam\cite{a19}. In the future, spam filters are expected to become more intelligent and able to distinguish safe emails from those that need to be removed from the inbox\cite{a20}. A new model based on deep learning algorithms has been developed for automatic spam detection and filtering\cite{a21,b10,c3}. Therefore, future work on spam detection could include further research and development of machine learning models that can accurately detect and filter spam while minimising false positives. There could also be a focus on developing more intelligent filters that can adapt to new types of spam attacks. It is also possible to identify and remove spam emails using the Mallet tool\cite{a22} and the DBPedia ontology\cite{a23}.
\bibliographystyle{IEEEtran}
\bibliography{References}
\end{document}